\PassOptionsToPackage{numbers,sort&compress}{natbib}

\documentclass{article}

\usepackage[final]{style/neurips_2024}

\usepackage[utf8]{inputenc}
\usepackage[T1]{fontenc}

\DeclareUnicodeCharacter{202F}{\,}   
\DeclareUnicodeCharacter{00A0}{~}    
\DeclareUnicodeCharacter{2013}{--}   
\DeclareUnicodeCharacter{2014}{---}  
\DeclareUnicodeCharacter{2212}{-}    
\DeclareUnicodeCharacter{00D7}{\(\times\)} 

\usepackage{hyperref}
\usepackage{url}

\usepackage{booktabs}
\usepackage{amsfonts}
\usepackage{amsmath}
\usepackage{nicefrac}
\usepackage{microtype}
\usepackage{xcolor}
\usepackage{dsfont}
\usepackage{derivative}
\usepackage[pdftex]{graphicx}

\usepackage{enumitem}

\renewcommand{\mathbb}{\mathds}

\title{DINOv3 as a Frozen Encoder for CRPS-Oriented Probabilistic Rainfall Nowcasting}

\author{%
  {\textbf{Luciano Araujo Dourado Filho$^{1,2}$\thanks{Corresponding author: lucianoadfilho@gmail.com}, Almir Moreira da Silva Neto$^{1}$ Anthony Miyaguchi$^{3}$,}}\\
  {\textbf{Rodrigo Pereira David$^{2}$, Rodrigo Tripodi Calumby$^{1}$, Lukáš Picek$^{4}$}}\\
  $^1$ Advanced Data Analysis and Management, University of Feira de Santana \\
  $^2$ National Institute of Metrology, Quality and Technology \\
  $^3$ Georgia Institute of Technology,
  $^4$ University of West Bohemia in Pilsen\\
}

\begin{document}
\maketitle

\begin{abstract}
This paper proposes a competitive and computationally efficient approach to probabilistic rainfall nowcasting. A video projector (V-JEPA Vision Transformer) associated to a lightweight probabilistic head is attached to a pre-trained satellite vision encoder (DINOv3\text{-}SAT493M) to map encoder tokens into a discrete empirical CDF (eCDF) over 4-hour accumulated rainfall. The projector-head is optimized end-to-end over the Ranked Probability Score (RPS). As an alternative, 3D-UNET baselines trained with an aggregate Rank Probability Score and a per-pixel Gamma-Hurdle objective are used. On the Weather4Cast 2025 benchmark, the proposed method achieved a promising performance, with a CRPS of 3.5102, which represents $\approx$26\% in effectiveness gain against the best 3D-UNET. 
\end{abstract}

\section{Introduction}

Short-term precipitation forecasting (``nowcasting'') underpins risk-aware decisions in urban drainage and flood mitigation, aviation safety, road traffic management, and renewable energy balancing. Urban flash-flood control and sewer operations depend on reliable exceedance probabilities at neighborhood scale; airlines and air-traffic controllers require accurate, rapidly updating assessments of convective activity; grid operators need short-horizon rain and cloud dynamics to schedule reserves and balance wind/solar variability \citep{schaible2024pvnowcasting}.
Meeting these operational needs requires models that deliver both high spatial/temporal resolution and calibrated uncertainty. Despite rapid progress, reliable high-resolution forecasts from heterogeneous observations remain challenging, particularly under spatial and temporal distribution shifts \citep{ravuri2021skillful,li2022w4c}.. Recent learning-based nowcasting systems—ranging from ConvLSTM video predictors to multi-resolution neural weather models—have advanced the state of the art at fine lead times \citep{shi2015convlstm,sonderby2020metnet,espeholt2022metnet2}. In parallel, probabilistic evaluation via proper scoring rules such as the Continuous Ranked Probability Score (CRPS) has become standard, rewarding calibrated predictive distributions rather than point estimates \citep{gneiting2007strictly,hersbach2000crps}.

We study the cumulative-rainfall track in a cross-sensor benchmark setting that stresses generalization. The target is the four-hour accumulation (16 slots at 15-minute cadence) averaged over a $32{\times}32$ radar window, while inputs are recent multi-band satellite frames. This configuration forces super-resolution from satellite to radar scale and exposes models to domain shifts in space and time. Prior work around the Weather4cast line of benchmarks has shown that formulating geospatial forecasting as video prediction yields strong baselines and controlled stress tests under spatiotemporal and domain distribution shifts. \citep{w4c24-whitepaper,belousov2022w4c}.

Our approach takes a pragmatic view of foundation models for Earth observation: reuse a large satellite-pretrained vision encoder as a frozen ``world-model’’ prior, and learn only a narrow probabilistic projector for downstream nowcasting. Concretely, we use a frozen DINOv3 ViT-L/16 encoder trained on satellite imagery \citep{dinov3} and compare lightweight transformer heads that map latent tokens to a categorical distribution over four-hour rainfall-accumulation bins. The probabilistic heads are trained with a discrete CRPS loss computed from cumulative probabilities (an eCDF), aligning optimization with the evaluation metric. To contextualize transfer versus full end-to-end learning, we also train compact, fully learnable 3D-UNet baselines that operate on the spatiotemporal satellite stack, instantiated with (i) a discrete-CRPS eCDF head and (ii) a zero-inflated Gamma (Gamma-Hurdle) head whose samples are aggregated into an empirical CDF of the four-hour total \citep{ronneberger2015unet,cicek2016unet3d}. This design allows a controlled comparison of (frozen encoder + narrow head) against a from-scratch spatiotemporal convolutional alternatives, under a scoring rule that foregrounds calibration.

\paragraph{Contributions.}
\begin{enumerate}[label=(\roman*), leftmargin=5mm, itemsep=0pt, topsep=2pt]
\item We present a focused case study of frozen DINOv3 features for precipitation nowcasting from satellite context.
\item We introduce a simple, calibration-oriented eCDF head trained with a discrete CRPS objective aligned to evaluation.
\item We benchmark against compact 3D-UNet variants with both eCDF and hurdle-Gamma probabilistic heads, clarifying the trade-offs between sample efficiency, parameter count, and probabilistic quality.
\end{enumerate}

\section{Data}

The dataset is comprised of two years of 11 band satellite image radiance (visible, water vapor, and infrared) and OPERA high resolution rain rates in mm/hr at quarter hour frequency.
The rain rates are six times the resolution of the satellite imagery meaning a 32x32 grid in radar resolution corresponds to an approximate 6x6 patch in the satellite resolution.

For each training sample we extract a co-registered satellite context (4 frames, 11 bands, last hour) padded with context to match the corresponding $32\times32$ radar window used by scoring.
This preserves the native super-resolution relationship between satellite context and radar target.
During training we sample the radar data-cube with rain-fall biased random crops and apply random rotation and reflection during training.
During validation, we center or evenly tile the cropping procedure for determinism.

\begin{table}[h!]
\centering
\caption{Summary Statistics of the Validation Data Split}
\label{tab:val_stats}
\begin{tabular}{lrrrrr}
\toprule
\textbf{Statistic} & \textbf{Mean} & \textbf{Min} & \textbf{p50} & \textbf{p95} & \textbf{Max} \\
\midrule
Masking (\%) & 0.0000 & 0.0000 & 0.0000 & 0.0000 & 0.0000 \\
Zero-Inflation (\%) & 88.8463 & 0.0916 & 99.8718 & 100.0000 & 100.0000 \\
Aggregate Target (mm) & 0.3434 & 0.0000 & 0.0012 & 2.4084 & 4.1663 \\
\midrule
Non-Zero Mean (mm) & 0.3073 & 0.0000 & 0.2354 & 0.9859 & 1.6961 \\
Non-Zero Median (mm) & 0.2327 & 0.0000 & 0.2200 & 0.7100 & 1.1700 \\
Non-Zero p95 (mm) & 0.7607 & 0.0000 & 0.2900 & 2.9515 & 4.7800 \\
Non-Zero Max (mm) & 1.7536 & 0.0000 & 0.2950 & 7.9700 & 10.5800 \\
\bottomrule
\end{tabular}
\end{table}

We compute aggregate statistics over the validation radar data cubes in Table \ref{tab:val_stats}, demonstrating a high degree of zero-inflation.
Rain fall in this split is rare, with a median per-sample median of 99.9\% zero-valued pixels.
The median per-sample target rain fall rate is 0.0012 mm/hr, while the non-zero-inflated value is 0.235mm/hr.
The challenge is then learning how to learn for a target signal that is averaged over a sparse spatio-temporal cube.

\subsection{Training Objective}
\begin{figure}
    \centering    \includegraphics[width=1.0\columnwidth]{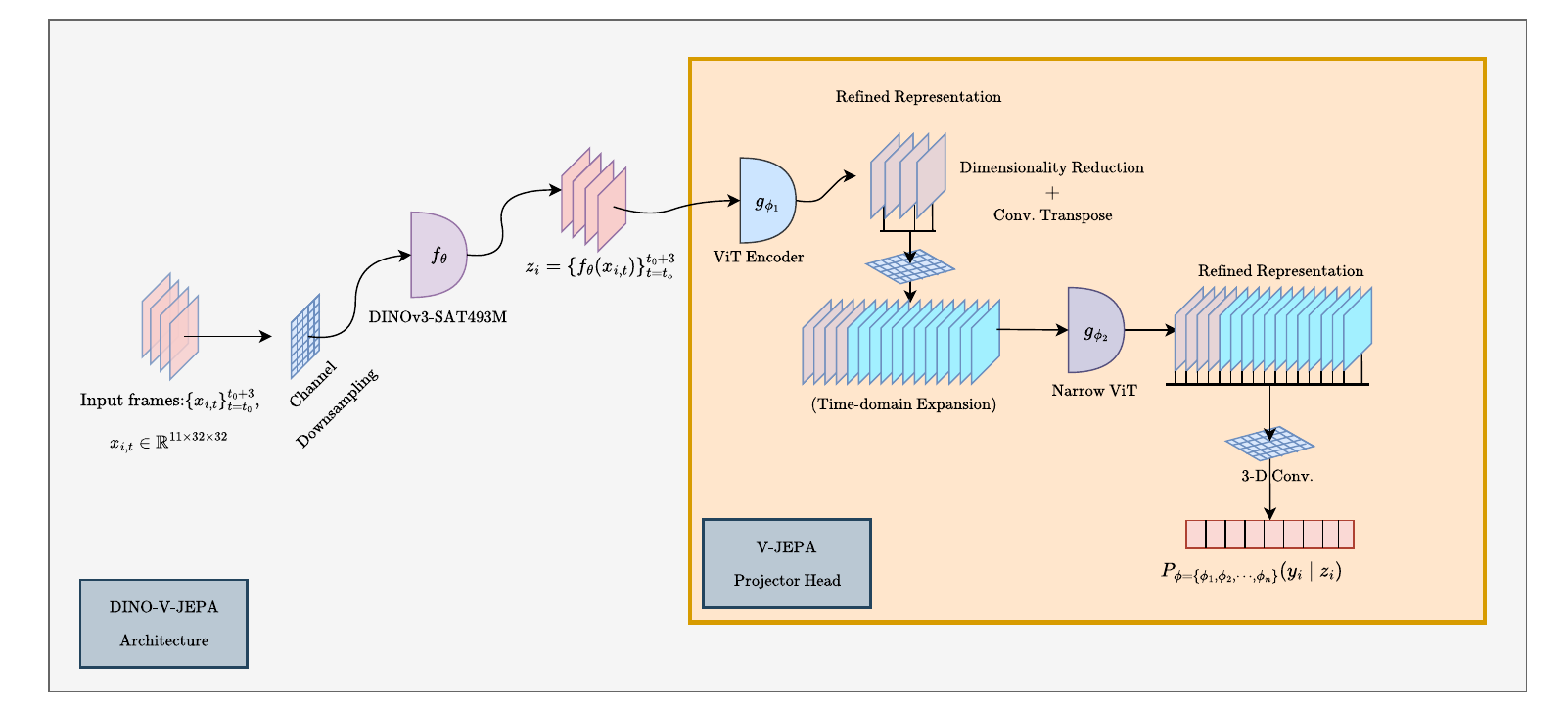}
    \caption{Training pipeline illustration for the DINO-V-JEPA architecture.}
    \label{fig:dino_vjepa}
\end{figure}
The target variable $y_{i,t}$, is defined as the average of ground-truth rain rate ($R_t \in \mathbb{R}$ ) for each pixel $(m,n)$ over a $32\times32\ (H\times W)$ grid, averaged within the t-th time-slot ($t \in [t_0, t_0 + 15]$).

\begin{equation}
     \label{eq:target_variable}
     y_i  = \frac{1}{4}\sum_{t=t_0}^{t_0+15} \left[\frac{1}{H\times W}\sum_{m=1}^{H}\sum_{n=1}^{W}R_{t,m,n} \right]
\end{equation}

We propose training the models over a discrete version of the Continuous Probability Rank Score (CRPS)~\cite{crps_paper} metric, namely Rank Probability Score (RPS), described in Equation~\ref{eq:rps}. 


\begin{equation}
    \label{eq:rps}
    RPS(F, y) = \sum_{k=1}^K \big(F_k - \mathds{1}(k \geq y) \big)^2
\end{equation}

Let $P_\phi(y_i \mid z_i)$ be the probability distribution of the target variable (spatially-averaged cumulative rainfall) conditioned on a latent representation of the input frames $z_i = \{f_\theta(x_{i, t}) \}_{t = t_o}^{t_0 + 3}$: the set of embeddings extracted for an arbitrary geographical region $i$ in each time instant, ranging from $t= t_0$ to $t_0 + 3$ -- where $t_0$ indicates an initial time-step. $f_\theta(x_{i,t})$ denotes the encoding function (e.g., 3D-UNET, DINOv3, V-JEPA, etc) that maps an input satellite frame $x_{i,t} \in \mathbb{R}^{11\times 32\times 32}$ into an arbitrary representation. Then $F_k$ (Equation~\ref{eq:f_sub_k}), represents the cumulative distribution function (CDF) at bin $k$, where $p_k$ denotes the predicted probability mass from $P_\phi(y \mid z)$ at bin $k$.  

\begin{equation}
  \label{eq:f_sub_k}
  F_k(P_{\phi}(y_i \mid z_i)) = \sum_{j=1}^k p_j
\end{equation}

The training objective with the standard RPS loss then amounts to find the set of model parameters $\phi$ from a $P_\phi(y \mid z)$ whose CDF minimizes the squared distance to the target CDF of rainfall values (Equation~\ref{eq:training_objective}). For the DINO-V-JEPA model (illustrated in Figure~\ref{fig:dino_vjepa}), for example, $P_\phi(y_i \mid z_i = \{f_\theta(x_{i, t}) \}_{t = t_o}^{t_0 + 3})$ is a categorical distribution over the $K$ bins, where $f_\theta$ represents the frozen DINOv3, and $\phi$ denotes the V-JEPA projector parameters. For the 3D-UNET variant, for example, $P_\phi(y_i \mid z_i)$ denotes the Gamma-Hurdle distribution parameterized by the per-pixel outputs of the model head whereas $\phi$ encompasses both the parameters of the feature extractor/encoder $f_\theta$ as the decoder, encoder and   probabilistic modeling parameters, as the UNETs were trained end-to-end.



 \begin{gather}
  \label{eq:training_objective}
  \mathcal{L_{RPS}} = \frac{1}{N} \sum_{i=1}^N \sum_{k=1}^K \bigg( F_k(P_{\phi}(y_i \mid z_i))  - \mathds{1}(k\geq y_i)\bigg)^2
 \end{gather}


\section{Proposed Architectures}

For this study, we assessed the spatio-temporal modeling inductive-biases for two architectural variants: Vision Transformers and 3D-UNET models. Some experiments covered complementary variants sharing the same probabilistic head and discrete-CRPS training objective in which the model outputs binned logits over rainfall amounts that are converted empirical CDFs.

\subsection{Vision transformer projector over a frozen DINOv3 backbone}
For the Vision Transformer models, we combined a pre-trained satellite encoder (DINOv3-SAT493M), as a frozen backbone to an asymmetrical encoder-decoder projector based on the V-JEPA~\cite{vjepa2} ViT implementation. A general illustration of the proposed architecture is depicted in the Figure~\ref{fig:dino_vjepa}. Analogously to~\cite{filho2025zero}, the rationale behind this variant (DINO-V-JEPA) is to harness the unknown capability of the pre-trained DINOv3 world model in providing a prior on the latent representation of satellite images that is conceivably informative enough to allow the projector to learn a conditional distribution of the accumulated rainfall. In other words, the projector role is to model time-dependent relationships between input frame embeddings $ \{f_\theta(x_{i, t}) \}_{t = t_o}^{t_0 + 3}$ extracted by the DINOv3 backbone, in such a way that it learns a predictive representation of cumulative rain. 

In order to achieve this, we had to account for the distribution-shift between three-channel (RGB) images (DINOv3 pre-training and input requirements) to 11-channel satellite images, which we approach by including a trainable downsampling layer to bridge the modality gap while retaining spectral information. This way, each of the 4 context frames random crops $\mathbf{x}\in\mathbb{R}^{11\times 4\times 32\times 32}$ is resized to a $224\times224$ pixel input resolution, and passed independently through the frozen (ViT--L/16) DINOv3 backbone for feature extraction. This allows obtaining the latent representation $\mathbf{z}\in\mathbb{R}^{4\times 196\times d}$ (per-frame patch tokens) with $d=1024$, that is passed into the DINO-V-JEPA projector. 

Architecturally-wise, the DINO-V-JEPA projector is an asymmetrical encoder-decoder ViT (V-JEPA~\cite{vjepa2} architecture) trained from scratch to address the task of modeling long-term (time-dependent) relationships between $\mathbf{z}\in\mathbb{R}^{4\times  196\times 1024}$ tokens (encoder) then projecting this representation towards a reduced dimensionality space, and mapping this reduced dimensionality representation back into the pixel space for cumulative rainfall prediction (decoder).    

More specifically, the encoder is a 8-layer ViT~\cite{vjepa2} that takes as input the DINOv3 embeddings $z_i = \{f_\theta(x_{i, t}) \in \mathbb{R}^{196\times 1024} \}_{t = t_o}^{t_0 + 3}$ and map them it into an unified latent representation $\mathbf{z}\in\mathbb{R}^{4\times196\times 1024}$. The decoder then projects $\mathbf{z}\in\mathbb{R}^{4\times  196\times 1024}$ into a compressed representation $\mathbf{z'}\in\mathbb{R}^{4\times  196\times 384}$ -- through a sequence of linear transformations (and non-linear activations). After compressing the encoder representation, the decoder then interpolates $\mathbf{z'}$ from 4 to 16 frames (i.e., $\in \mathbb{R}^{16 \times 196 \times 384}$) via learnable transformation (2-D transposed convolution operator). This interpolated representation is fed to a narrow ViT (6 transformer blocks) and then through a 3-D convolution for collapsing spatio-temporal features into the $K$ rainfall bins. We believe that performing a ``time-series’’ expansion from 4 into 16 frames conceivably generates a structured representation that provide a more appropriate inductive bias towards facilitating the prediction of the accumulated rainfall over the K bins. 

Although we use the V-JEPA ViT implementation as a projector, this model is trained from scratch (in a supervised fashion) to match the target CDFs -- according to the objective illustrated in Equation~\ref{eq:training_objective}. The model outputs are discretized into $K$ bins, from which we establish an upper bound $R_{max}$ (mm/hr) and a bin width $\epsilon$, which allows us to define $K =\left[\frac{R_{max}}{\epsilon}\right]+1$ bins.

\subsection{3D-UNET with Probabilistic Hurdle Head}

We experiment with a 3D-UNET model as a feature extractor with a probabilistic head for regularization for the problem.
A 3D-UNET has the inductive-bias for the spatio-temporal modeling necessary for the task.
The banded imagery and raster product is passed through the UNET encoder, which is upscaled and smoothed via trilinear filtering from 4 frames to 16 frames and refined via a (2+1)D convolutional layer.
We apply several variants on the head.

\subsubsection{Aggregate Rainfall via RPS Loss}

The first variant of this model predicts the discretized rank probability score directly to the averged prediction for the cell.
The encoder is fed into a multi-head attention pooling layer which then predicts logits for each bin.

\subsubsection{Per-Pixel Rainfall via Gamma-Hurdle Loss}

This loss jointly optimizes the Negative Log-Likelihood (NLL) of the hurdle and Gamma components by learning their relative uncertainty.
Let $R$ be the true rainfall at a given pixel, and $(l, \alpha, \beta)$ be the predicted parameters for that pixel. Let $\mathbb{1}_V$ be the indicator function for a valid pixel.
The per-pixel hurdle loss $L_h$ (implemented as a Focal Loss) and the Gamma Negative Log-Likelihood (NLL) $L_g$ are defined as:

\begin{equation}
\label{eq:pixel_losses}
L_h = \text{FocalLoss}(l, \mathbb{1}_{R > 0}), \quad L_g = \mathbb{1}_{R > 0} \cdot \left[ -\log(\text{Gamma}(R | \alpha, \beta)) \right]
\end{equation}

Let $G_{\sigma_b}$ be a 2D Gaussian blur kernel with standard deviation $\sigma_b$. 
These losses are smoothed spatially at each time step (where $*$ denotes a convolution over spatial dimensions):

\begin{equation}
\label{eq:spatial_smoothing}
\tilde{L}_h = L_h * G_{\sigma_b}, \quad \tilde{L}_g = L_g * G_{\sigma_b}
\end{equation}

Using learnable log-variances $s_h = \log \sigma_h^2$ and $s_g = \log \sigma_g^2$ for multi-task weighting with uncertainty \cite{kendall_multi-task_2018}, the combined loss $\mathcal{L}_{\text{pixel}}$ is the average of the valid ($\mathbb{1}_V$), smoothed, per-pixel weighted losses:

\begin{equation}
\label{eq:combined_loss}
\mathcal{L}_{\text{pixel}} = \frac{\sum \mathbb{1}_V \cdot \left( \left[ e^{-s_h} \tilde{L}_h + \frac{1}{2}s_h \right] + \left[ e^{-s_g} \tilde{L}_g + \frac{1}{2}s_g \right] \right)}{\sum \mathbb{1}_V + \epsilon}
\end{equation}

An empirical CDF is formed by sampling rainfall means computed by the expected value of the hurdle-gamma distribution, which is the product of the hurdle probability and the mean of the Gamma component:

\begin{equation}
R_\mu = \text{sigmoid}(l) \cdot \left(\frac{\alpha}{\beta}\right)
\end{equation}

These samples are aggregated and binned appropriately for the task.
We may also directly predict a point prediction using mean across parameters in a cell.

\subsubsection{Multi-Task Aggregate and Per-Pixel Loss}

The last head is a combination of the aggregate and per-pixel heads.
We optimize for both heads using the aggregate RPS and Gamma-Hurdle losses together using an Exponential Moving Average (EMA) Loss Weighting strategy \cite{lakkapragada_mitigating_2022}.
The per-pixel and aggregate losses are normalized by their moving average to map onto a comparable scale before being averaged together.
Normalization is necessary because the aggregate loss is in physical units of mm/hr while the probabilistic loss is a negative log likelihood, and otherwise results in ill-posed gradients. 

\begin{equation}
\mathcal{L}_{\text{total}} = \frac{\mathcal{L}_{\text{pixel}}}{\tilde{\mathcal{L}}_{\text{pixel}} + \epsilon} + \lambda_{\text{agg}} \cdot \frac{\mathcal{L}_{\text{agg}}}{\tilde{\mathcal{L}}_{\text{agg}} + \epsilon}
\label{eq:autobalance_loss}
\end{equation}

We treat the per-pixel head as a form of regularization and use only the predicted eCDF bins.

\section{Results}

Submissions to the Weather4cast competition are summarized in Table~\ref{tab:model_scores}. These results represents the CRPS at the test dataset obtained for each corresponding modeling strategy. As observed, the DINO-VJEPA model, trained over the standard RPS objective with $R_{max} =128(mm/hr)$ achieved the best performance with CRPS=$3.5102$. With respect to the 3D-UNET models, we observed that the probabilistic per-pixel head does marginally better than an equivalent model that directly predicts the eCDF. In other words, solving for the multi-task objective does no better than the RPS objective alone. Using the wider bins in the v4 model effectively reduces the task down to binary classification, as the distribution of predicted scores concentrated towards either 0mm/hr or 4mm/hr.

\begin{table}[h]
\caption{Summary of model scores and parameters.}
\label{tab:model_scores}
\centering
\begin{tabular}{l l c c l}
\toprule
Backbone & Loss Objective & Bin Max (mm/hr) & Num Bins (K) & CRPS \\
\midrule
DINO-VJEPA & RPS (Aggregate) & 128 & 25601 & 3.5102 \\
UNET3D (v4) & Gamma-Hurdle (Per-Pixel) & 512 & 129 & 4.7637 \\
DINO-VJEPA & RPS (Aggregate) & 128 & 129 & 5.5894 \\
UNET3D (v10) & Gamma-Hurdle (Per-Pixel) & 64 & 6401 & 6.3634 \\
UNET3D (v10) & RPS (Aggregate) & 64 & 6401 & 7.0249 \\
UNET3D (v10) & RPS + Gamma-Hurdle & 64 & 6401 & 7.1057 \\
\bottomrule
\end{tabular}
\end{table}

Table~\ref{tab:dino_vjepa_scores} presents a detailed view on the per-epoch results for the DINO-VJEPA models. For this class of models we observed two general tendencies with regard the CRPS results on the test set. The first concerns bin width/granularity, from which as opposed to the 3D-UNET models, displayed a far more significant role for the DINO-VJEPA variants. Secondly, we observed that by maintaining $R_{max}$ fixed at 128 (mm/hr), increasing the amount of bins not only significantly improved the model CRPS while increased training stability, yet early-stopping was still mandatory to overcome training instability, as the models haven't demonstrated signals of convergence.  

\begin{table}[h]
  \caption{Results for the DINO-VJEPA mapping from 0-128(mm/hr), with $K=\{129, 25601\}$}.
  \label{tab:dino_vjepa_scores}
  \centering
  \begin{tabular}{lll|lll}
    \toprule
    Epoch     & $K$     & CRPS & Epoch     & $K$     & CRPS \\
    \midrule
     1 & 129 &  5.6870 & 1 & 25601 & 4.9539\\
     2 & 129 &  5.9988 & 2 & 25601 & 4.7817\\
     3 & 129 &  6.2564 & 3 & 25601 & 4.5317\\
     4 & 129 &  5.5894 & 4 & 25601 & 3.9821\\
     5 & 129 &  6.2577 & 5 & 25601 & 3.6404\\
     6 & 129 &  6.2810 & \textbf{6} & \textbf{25601} & \textbf{3.5102} \\
     7 & 129 &  5.8653 & 7 & 25601 & 6.6455 \\
     8 & 129 &  6.2877 & 8 & 25601 & 5.5621  \\
     9 & 129 &  6.3051 & 9 & 25601 & 6.7276  \\
     10 & 129 & 6.1426    & 10 & 25601 & 5.8937  \\
    \bottomrule
  \end{tabular}
\end{table}

\section{Discussion}

Our results points towards practical lessons for probabilistic nowcasting from satellite images. In overall, one of the most decisive modeling aspects identified by us concerned criteria related to rainfall i.e., max rain ($R_{max}$) and number of bins (K). Even though the dataset distribution demonstrated to be heavily-skewed (with e.g., 0.235 mm/hr median--  Table~\ref{tab:val_stats}), the models demonstrated to benefit from a $R_{max} \geq 128\ (mm/hr)$, which not only evidences the discrepancy between the train/val split but the importance of accounting for extreme values (e.g., $\geq 32\ (mm/hr)$~\cite{w4c24-whitepaper}. 

For the DINO-V-JEPA model variants, for example, bin granularity $\epsilon$, demonstrated to be one of the most decisive aspects. We hypothesize that these variants, in contrast to UNET models, benefited more on fine-grained rainfall modelling rather than a coarse-grained set of accumulation bins, conceivably due to the amount of adjustable parameters. We believe that this happens because when rainfall is represented with a coarse set of accumulation bins ($K=129$), the model’s eCDF can only snap to a few thresholds, which conceivably not only induces discretization bias (by masking small but meteorologically meaningful changes in accumulation), but facilitates overfitting. Using fine-grained bins ($K=25601$), on the other hand seemed to reduce that bias, allegedly by sharpening the learning signal (as the RPS gradient builds upon residuals at each threshold), but more decisively perhaps, making the learning objective harder, as we propose to model a much more fine-grained probability distribution. 

In addition to that, we highlight the role of the frozen, domain-aligned backbone as an efficient/informative prior. Our hypothesis is that treating a pre-trained satellite-encoder as a fixed ``world-model'' with a narrow projector on top, seemed to further concentrate the optimization on the probabilistic readout and feature refinement (e.g., time-domain alignment), rather than representation learning, which empirically leads to faster convergence, and well-calibrated eCDFs. On the other hand, overfitting seemed to set in quickly, conceivably once the projector head saturates the information from the frozen features, making early-stopping mandatory in the absence of more effective regularizers--- a training dynamic also observed in related domains with frozen DINO encoders and small task heads~\cite{filho2025zero} which emphasizes research questions for future work. 


\subsection{Probabilistic Modeling}

Probabilistic modeling can effectively model zero-inflation of the data, often performing better than some of the direct counterparts.
The hurdle aspect of the model contributes most heavily to performance, as we see in our results that a de-facto two-bin model with a max of 4mm/hr outperforms other models with more flexibility.
We experiment non-standard techniques like Gaussian blurring over backbone outputs in order to introduce a spatial bias, instead of explicit modeling like Gaussian Processes or Conditional Autoregression.
The focal loss in particular helps significantly over a sigmoid associated with Bernoulli variables, as it can adapt to the extreme skew.
However, due to the fact that we need to predict a single aggregated target, we lose out on a lot of information that we could otherwise be exploited by explicit modeling.
If we were tasked with estimating the distribution of rainfall (e.g. histograms instead of a mean), this approach may have been aligned with the necessary inductive bias.

One alternative to the RPS loss that we tried was taking the per-pixel rainfall parameters and averaging them to fit the moments of a new distribution.
Our distribution should be strictly positive for estimating a physical quantity, parameterized analytically from the mean and variance, and have a differentiable CDF. 
We experiment with the Log-Normal and Gamma distributions.

\begin{equation}
\label{eq:moment_matching}
\sigma^2_{\text{norm}} = \log\left(\frac{\text{Var}(y)}{\max(E[y]^2, \epsilon)} + 1\right), \quad \mu_{\text{norm}} = \log(\max(E[y], \epsilon)) - \frac{\sigma^2_{\text{norm}}}{2}
\end{equation}


When we try to optimize for the multi-task objective using the aggregate moment matching process with the per-pixel objective, we find that the gradient is much harder to achieve with the log-normal distribution using uncertainty weighting.
While theoretically sound, this causes issues when we try to create a hybrid between the per-pixel rainfall rates and the aggregate rainfall rate because the resulting average is pushed strongly toward very small values due to zero inflation and performed poorly.

For the log-normal distribution, we face a numerical stability issue when trying to propagate the gradient through the loss of the analytical CDF to the empirical CDF.
The fitted Gamma shape parameters pass through a log function, which causes the gradients to explode when the values are near zero. 
The gamma distribution is numerically stable in comparison, because it only relies on divisions and multiplications. 
However, we find that the CDF of the gamma distribution in PyTorch is not differentiable with respect to the first argument (as it involves computation of the Meijer G-function). 
We can use the reparameterization trick to instead sample from the distribution and have gradients flow backwards from a soft-binned CDF.
Due to complications with exploding gradients, we gave up multi-task probabilistic objectives. 


\section{Conclusion}
This work presented an extensive experimental study of probabilistic rainfall nowcasting using both transformer and convolutional-based architectures. Two model families were assessed: (i) a lightweight probabilistic projector head attached to a frozen DINOv3\text{-}SAT493M encoder (DINO\text{-}V\text{-}JEPA), and (ii) 3D\text{-}UNET backbones with several probabilistic output heads. Among all configurations, the DINO\text{-}V\text{-}JEPA projector with fine discretization (0–128\,mm, $K{=}25{,}601$) achieved the best test performance, reaching a \textbf{CRPS of 3.5102}. 

Fine-grained discretizations improved early convergence but also led to instability in later epochs, indicating sensitivity to binning resolution and optimization dynamics. Within the UNet family, the per-pixel Gamma–Hurdle head achieved the strongest performance (\textbf{CRPS = 4.763743}), while aggregate eCDF and multi-task objectives did not provide further improvements. These experiments demonstrate that pretrained satellite representations can be effectively reused in a frozen state, with small probabilistic heads delivering competitive empirical CDF forecasts at a fraction of the computational cost required by fully trainable baselines. Overall, our results support the idea that pretrained world models, combined with simple probabilistic heads, are a promising and practical direction for operational nowcasting under strong sparsity and distribution-shift.



\paragraph{Limitations and Future Work}
Our analysis is limited to a single benchmark track (4-hour cumulative rainfall) and a small set of architectural and binning choices, therefore some findings might not be reflective or generalizable enough for straightforward domain transfer. We did not systematically explore extensive discretizations, regularization schemes, or parameter-efficient fine-tuning of the frozen encoder, besides that, multi-task probabilistic objectives were only tested in a few numerically fragile variants. Future work should study the interaction between bin granularity and optimization in a more controlled way, evaluate the approach across regions and tasks, and extend the framework to richer inputs and multiple accumulation horizons, while also reporting a more detailed computational cost profile.

Code for all 3D\text{-}UNET baselines and probabilistic heads is publicly available at \url{https://github.com/acmiyaguchi/weather4cast-2025}. Code and parameters for the DINO-VJEPA models are available at \url{https://github.com/FalsoMoralista/Weather-4-Cast}. 

\section*{Acknowledgements}

This research was supported in part through research cyberinfrastructure resources and services provided by the Partnership for an Advanced Computing Environment (PACE) at the Georgia Institute of Technology, Atlanta, Georgia, USA \href{http://www.pace.gatech.edu}{(http://www.pace.gatech.edu)}, UEFS-AUXPPG 2023/2024/2025, CAPES-PROAP 2023/2024/2025,
CAPES grant 88887.159255/2025-00 and 88887.594676/2020-00 and UEFS FINAPESQ (grant 047/2023). 

\bibliographystyle{unsrtnat}   
\bibliography{references}      
\end{document}